\documentclass{article}

\usepackage[english]{babel}
\usepackage{hyphenat}
\usepackage{microtype}
\usepackage{tabularx}
\usepackage{arxiv}

\usepackage{makecell}

\usepackage[T1]{fontenc}
\usepackage{url}  
\usepackage{booktabs}    
\usepackage{amsfonts}    
\usepackage{nicefrac}       
\usepackage{microtype}   
\usepackage{lipsum}	
\usepackage{graphicx}
\usepackage[numbers]{natbib} 
\usepackage{doi}

\usepackage{array}
\usepackage{xcolor}
\definecolor{darkgreen}{rgb}{0, 0.5, 0}

\usepackage{authblk}
\usepackage{titlesec}

\usepackage{hyperref} 

\newcommand{\authorsize}{\large}

\title{zGAN: An Outlier-Focused Generative Adversarial Network for Realistic Synthetic Data Generation}

\author[1]{\authorsize Azizjon Azimi}
\author[1]{\authorsize Bonu Boboeva}
\author[1]{\authorsize Ilyas Varshavskiy}
\author[1]{\authorsize Shuhrat Khalilbekov}
\author[1]{\authorsize Akhlitdin Nizamitdinov}
\author[1]{\authorsize Najima Noyoftova}
\author[1]{\authorsize Sergey Shulgin}

\affil[1]{zypl.ai: \texttt{azizjon@zypl.ai}; \texttt{bonu@zypl.ai}; \texttt{ilyas.varshavskiy@zypl.ai}; \texttt{shuhrat.khalilbekov@zypl.ai}; \texttt{\hspace*{0.3em}akhlitdin@zypl.ai}; \texttt{najima@zypl.ai}; \texttt{sergey.shulgin@zypl.ai}}

\renewcommand{\shorttitle}{\textit{arXiv} zGAN}

\hypersetup{
pdftitle={zGAN: An Outlier-Focused Generative Adversarial Network for Realistic Synthetic Data Generation},
pdfauthor={Azizjon Azimi, Bonu Boboeva, Ilyas Varshavskiy, Shuhrat Khalilbekov, Akhlitdin Nizamitdinov, Najima Noyoftova, Sergey Shulgin},
pdfkeywords={gan, outliers, synthetic data generation, tabular data},
}

\begin{document}
\maketitle

\setcounter{figure}{0}
\setcounter{table}{0}

\begin{abstract}
The phenomenon of "black swans" has posed a fundamental challenge to performance of classical machine learning models. The perceived rise in frequency of outlier conditions, especially in post-pandemic environment, has necessitated exploration of synthetic data as a complement to real data in model training. This article provides a general overview and experimental investigation of the zGAN model architecture developed for the purpose of generating synthetic tabular data with outlier characteristics. The model is put to test in binary classification environments and shows promising results on realistic synthetic data generation, as well as uplift capabilities vis-à-vis model performance. A distinctive feature of zGAN is its enhanced correlation capability between features in the generated data, replicating correlations of features in real training data. Furthermore, crucial is the ability of zGAN to generate outliers based on covariance of real data or synthetically generated covariances. This approach to outlier generation enables modeling of complex economic events and augmentation of outliers for tasks such as training predictive models and detecting, processing or removing outliers. Experiments and comparative analyses as part of this study were conducted on both private (credit risk in financial services) and public datasets.\end{abstract}


\section{Introduction}

The exponential increase in data collection capabilities has accelerated the availability of data for machine learning training. Nonetheless, concurrent rise in regulatory measures surrounding data accessibility has furthered the challenge of data insufficiency. A primary factor contributing to this challenge is data privacy, which may include personal identifiable information, trade secrets, or intellectual property-related data \cite{Assefa2021, Shokri2017}.

Another significant factor contributing to data insufficiency is the periodicity of processes in economic, epidemiological, and social spheres. Examples of such processes include macro events such as crises or pandemics that result in significant socioeconomic repercussions \cite{Bluwstein2023, Ghil2011, Ajagbe2024}. Such processes are characterized by low predictability due to inherent uncertainty and lead to \textit{a posteriori} incorporation into model training \cite{Wabartha2020}.

Over the past decade, the widespread use of generative models has become more common to tackle data availability issues – with rapid emergence of Generative Adversarial Networks (GANs) \cite{goodfellow2014}. GANs are designed to generate synthetic data with resemblance to real historical data \cite{Bourou2021}. The design of GAN architecture embeds an assumption that the distribution of generated synthetic data ought to replicate the distribution of historical training data.

Hence, given the rise in outlier data samples that constantly challenge the assumptions of historical datasets, a need for augmentation of GANs for the purpose of adding utility and information value with respect to outliers has arisen. The zGAN model originally developed by zypl.ai\footnote{https://zypl.ai/en} aims to generate realistic synthetic tabular data with outlier characteristics to complement the information value of historical training data.

The generation of synthetic outliers by zGAN is intended to improve the predictability of rare events and to model fundamentally new events for further analysis. Generating outliers also allows augmenting existing datasets to enhance model training stability and train models capable of detecting, removing or processing outliers \cite{Nayak2024}.

Using zGAN, outliers can be generated in selected columns of datasets based on covariance matrices of real data derived from various customizable probability distributions with a specified limit on the distribution tails. The percentage of outliers relative to the total data in the selected columns is adjustable depending on the specific context of data samples. This feature of zGAN is grounded in the concept of Extreme Value Theory (EVT) and allows for the modeling of rare events across light, bounded, and heavy distribution tails \cite{deHaan2006}.

The realism of synthetic data generated by zGAN is ensured by the model’s enhanced ability to reproduce and maintain correlations of real training data within synthetically generated data sample. Privacy in zGAN is ensured by a similarity filter which maintains the confidentiality of generated data through the use of hash codes, allowing the model to not retain real client data post-training.

This article presents experimental results on generating synthetic data with respect to both private and open datasets. The experiments utilized multimodal tabular data containing numerical, temporal and categorical features. The quality of the generated synthetic data and model robustness were assessed using the "Area Under the Curve" (AUC) metric in binary classification experiments. Based on the correlation analysis of original and synthetic datasets, the ability of GANs to preserve and reproduce feature correlations is demonstrated.

The following are key scientific articles that were crucial in development of zGAN.

\paragraph{Related Works} SAGAN \cite{Zhang2018} introduces self-attention mechanisms to GANs, allowing the model to focus on different regions of the input space. This approach has significantly improved the quality of generated images by capturing long-range dependencies.

DA-GAN \cite{Yang2021} uses a dual attention mechanism, combining self-attention and channel-attention, to generate high-quality images. This model has shown improved performance in generating detailed and coherent images.

SA-CapsGAN \cite{Sun2021} incorporates a self-attention mechanism into a Capsule Network within GAN architecture. SA-CapsGAN has shown efficiency in modeling spatial feature relationships and capturing long-range dependencies.

HSGAN \cite{Li2021} is designed for 3D shape generation and uses a hierarchical structure with a self-attention mechanism. This architecture allows for capturing global graph topology in shape generation.

The paper is organized as follows: Section~\ref{sec:review} provides an overview of the latest scientific publications related to the subject of this work; Section~\ref{sec:architecture} describes the zGAN architecture; Section~\ref{sec:experiment_setup} discusses the experimental setup for each task, including a description of data samples and models used in the experiments. Section~\ref{sec:experimental_results} presents experimental results and their interpretation.

\section{Literature Review}
\label{sec:review}

This review summarizes the latest trends in areas related to generation of synthetic tabular data. The motivation for generating realistic synthetic tabular data and methods for calculating the similarity between real and synthetic data in the financial sector are discussed in \cite{Assefa2021}. The phenomenon of "Black Swans," which motivates the concept of zGAN, is described in the works \cite{taleb_black_2008, taleb2012antifragile}.

Self-attention algorithms are often used to improve model quality metrics. In studies \cite{Zhang2021, Oh2021}, researchers apply self- attention layers in GANs in combination with RNNs to enhance the correlation of data. The authors of PSA-GAN \cite{Paul2021} emphasize the importance of data correlation in experiments with time series and other tabular data \cite{Xu2018}. Self-attention layers are an integral part of DP-SACTGAN designed for tabular data \cite{Li2023_2}.

The use of self-attention layers in conjunction with transformers have also been explored \cite{Kossen2021, Somepalli2021}. In the study \cite{Mrini2019}, a Label-Attention layer was developed – allowing the application of self-attention algorithms to syntactic expressions. Using the network self-attention algorithm, it is possible to rank features of tabular datasets \cite{Skrlj2020}. The importance of correlation of tabular data is noted in the study on DP-CGANS, where experiments on imbalanced data show performance and privacy metrics of generated data \cite{Sun2022}.

To improve the performance of tabular GANs, including data correlation \cite{Vu2024}, researchers often modify the loss functions of models – including for CTGAN \cite{Pavlov2024} and FCT-GAN \cite{Zhao2022}. In the field of aerospace, a modified loss function is applied to the task of smoothing driving lines \cite{Chattoraj2024}.

The importance of information contained in outliers of tabular data is highlighted in the study \cite{Fallahian2024}. Concurrently, in the task of ensuring data security, outliers were found to contribute to re-identification of users \cite{Trindade2024}.

Detection and removal of outliers is a major topic in the context of anomalies and outliers in tabular data \cite{Du2024, Qi2024, Fang2024}. In the study \cite{Oh2019} dedicated to OD-GAN, experiments on the detection and removal of outliers are proposed. The  procedure of outlier removal and processing within tabular bankruptcy data is proposed in the article \cite{Nayak2024}.

The generation and detection of outliers in the context of financial reporting is proposed in the study \cite{Aftabi2023} while the same in the context of data augmentation for time series training is proposed in the work \cite{Zhao2024}.

The method for generating anomalies using a generator and discriminator, as well as subsequent binary classification for detecting irregularities in various applications is demonstrated in article \cite{pourreza2020}. In article \cite{Wabartha2020}, an ensemble approach called DENN is proposed for forecasting rare and unexpected events. The result is achieved by customizing the loss function which enhances the ensemble models’ ability to handle events that fall outside of training data distribution.

In the article \cite{Zakharov2024}, TRGAN demonstrates significant improvements in generating synthetic transactional data, particularly in maintaining accurate time intervals and outperforming other methods such as Banksformer, CTGAN, and CopulaGAN across various metrics. This highlights TRGAN’s robustness in handling both categorical and numerical attributes effectively.

\section{Architecture of zGAN}
\label{sec:architecture}
\begin{figure}[!htbp]
    \centering
    \includegraphics[width=\linewidth]{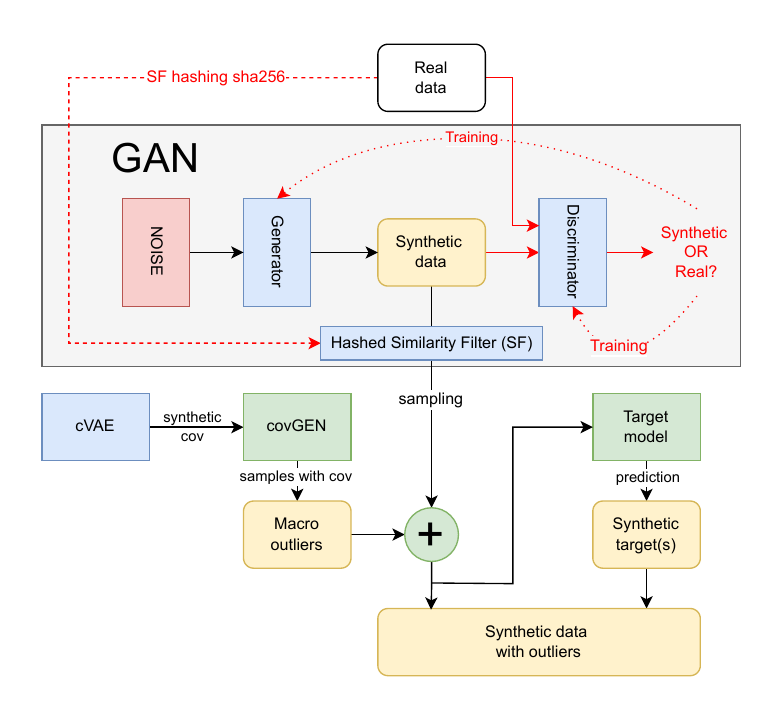}
    \caption{Generalized structure of zGAN.}
    \label{fig:zGAN_structure}
\end{figure}

The generalized structure of zGAN is shown in Figure \ref{fig:zGAN_structure}.

The generator of a Generative Adversarial Network (GAN) is initialized with random noise, producing synthetic data at its output. The discriminator receives both generated synthetic data and real data as inputs and outputs a binary code that corresponds to the "real" or "synthetic" data classes.

The training process of the discriminator involves enhancing its ability to distinguish synthetic data from real data, while the generator, on the contrary, is trained to produce data that is indistinguishable from real data at the discriminator’s output.

The privacy of synthetic data is ensured through the use of a similarity filter which employs hashes of real data to filter out generated synthetic data resembling real client data. This hash-based approach ensures that real client data is not retained after model training.

A GAN trained on real tabular data generates synthetic data that is indistinguishable from real data, into which macro outliers are integrated. The module that integrates macro outliers into the synthetic data is denoted by “\texttt{+}”.

Macro outliers are generated by the Outliers Conditional Covariance Generator (covGEN) through sampling from a multivariate manually fitted distribution, requiring either the covariance obtained from real data or synthetic covariance matrices generated by the Conditional Variational Autoencoder (cVAE). For generating outliers in accordance with EVT, the currently available distributions are normal, Laplace, Weibull, Gumbel, and Levy distributions.

cVAE generates synthetic covariance matrices corresponding to the structure of real data. The correspondence of the generated covariance matrices to a specific structure is achieved by minimizing the reconstruction error and the Kullback–Leibler divergence between the latent space distribution and the prior distribution of real data. The optimization problem is solved using a decoder network to reconstruct real data from the latent space, while the variational autoencoder learns representations of data in the latent space.

Synthetic tabular data may be necessary for tasks related to further analysis of certain dataset columns – targets \cite{Somepalli2021, Skrlj2020, Zhao2024}. For predicting target features integrated into synthetic data with outliers, the Target Model module based on a classification model, such as Catboost\footnote{https://github.com/catboost/catboost} is employed. The classification model is trained on a sample of real data to gather weak explanatory information from the dataset columns containing macro and micro data.

The ability of zGAN to generate realistic synthetic data can be experimentally verified and compared with the quality metrics of other GANs.

\section{Experiment Setup}
\label{sec:experiment_setup}

The proposed experiments demonstrate the ability of zGAN to generate realistic synthetic data in comparison with other GANs along with enhanced model robustness. The experiments are based on a classification task where the presence of a target variable in tabular data is used to compute AUC metric that characterizes the performance of classification task when comparing real and synthetic data.

Solving the classification task with a target feature generated together with other features allows assessing the ability of zGAN to generate synthetic data while preserving the feature relationships of the training dataset. For this purpose, baseline models trained just on real data are included in experimental analyses and AUC computation.

The classification task was performed in two versions: by splitting the data into training and test sets based on individual observations, i.e., "out-of-sample" (OOS), and by splitting the data based on time, i.e., "out-of-time" (OOT). The ability of zGAN to maintain the magnitude and variance of AUC when performing classification with different proportions of synthetic and real data, while reproducing the classification experiment on different datasets in OOS mode or on future data in OOT mode, characterizes the model’s robustness.

To demonstrate the ability of GANs to preserve and reproduce feature correlations, an analysis on correlation was conducted on the original and synthetic datasets.

\subsection{Classification task}

The methodology of conducted experiments implies equal conditions and the same sequence of operations for each dataset. The main stages of the conducted experiments are listed as follows:

\begin{itemize}
    \item Parsing Data;
    \item Data Preprocessing;
    \item Feature Selection;
    \item Splitting Data into Train and Test Sets;
    \item Baseline AUC Calculation;
    \item Train Data Sampling (optional);
    \item Fitting GAN on Real Train Data;
    \item Generating Synthetic Data;
    \item Sample of 80\% of the training and test data (optional);
    \item Fitting the Classifier on Synthetic Data;
    \item Calculating AUC on Test Data.
\end{itemize}

\begin{figure}[!htbp]
    \centering
    \includegraphics[width=\linewidth]{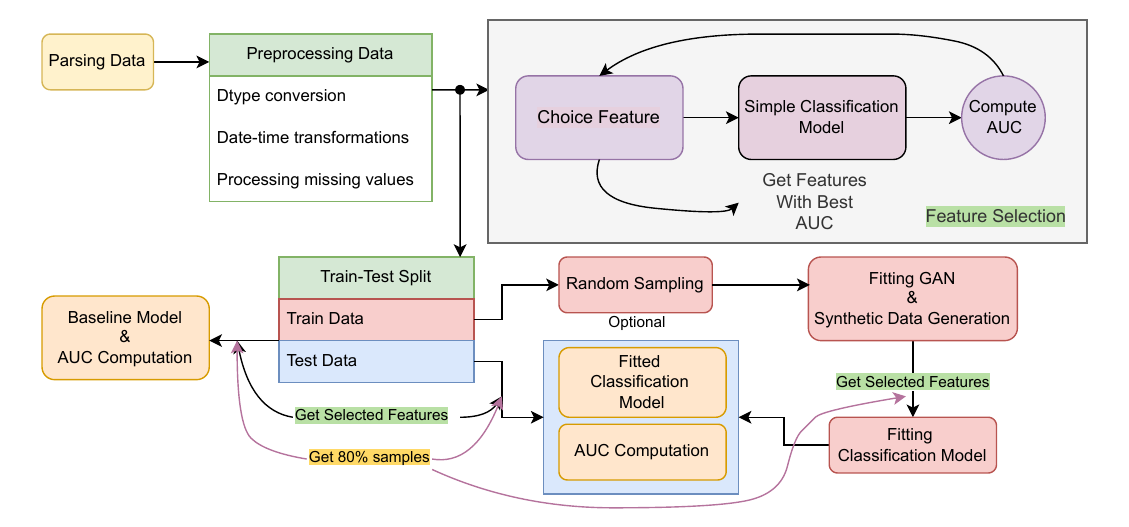}
    \caption{Experimentation pipeline.}
    \label{fig:experimentation_pipeline}
\end{figure}

A detailed description of conducted experiments is shown in the Figure~\ref{fig:experimentation_pipeline} architecture.

Data preprocessing was preceded by parsing. Preprocessing included data type conversions and date and time transformations when time series were present in the dataset. For models whereby fitting was not feasible with missing values, methods were applied to fill in the missing values.

Feature selection was based on analysis aimed at preventing data redundancy, data leakage, and informational noise. Features containing macro data were not used in the experiments. The primary goal of feature selection was to improve the performance of classification model. When training and generating data, all features presented in the datasets were used while for solving the classification task, only selected features were retained in the datasets.

\paragraph{Out-of-Sample experiments} In the OOS experiments, when splitting data into train and test sets, identical distribution laws within both parts of the original dataset were applied.

CatBoost was used for baseline models with quasi-optimal hyperparameter search to reduce overfitting. Grid search was utilized for hyperparameter tuning, with the grid search boundaries defined using random search.

The baseline classification model was fitted on the train data with AUC being calculated on the test data. The variability range of AUC was determined as follows: a sample constituting 80\% of the total data was randomly drawn from the training and test data separately. A classification model was then trained on this sample, and the resulting metric was calculated. The final metric presented in the study is the median AUC value obtained over the course of 51 experiments for each model and dataset separately. At the beginning of each experiment, a new sample was drawn.
 
The 51st value in the experimental results was the AUC value obtained without subsampling 80\% of the dataset, meaning it was calculated using all records from both training and test datasets.

Before fitting the synthetic data generators, for models sensitive to the amount of training data the dataset was augmented by random sampling. For models in which synthetic data realism was significantly worse than other models, manual hyperparameter tuning was applied. Manual tuning preferred configurations that achieved maximum AUC. In all cases, trained generative models and GANs were used to generate 4000 records each.

For final AUC calculation, the classification model was trained on synthetic training data and the performance was evaluated on real test data. The AUC variability in this case was calculated using a procedure entirely analogous to that described for the baseline model. The only difference is that the folds were created from synthetic training data rather than real training data.

In all experiments, the task of binary classification was addressed.

\paragraph{Out-of-Time experiments} In OOT experiments, the dataset was split into training and test sets based on an ascending order of dates. The data was split into training and test sets in two proportions: 50\% and 80\%.

In a classification experiment using OOT, a zGAN, a CTGAN and a classification model were trained on the real training dataset, and the AUC metric was calculated only on the real test dataset.

For further stability analysis, the training set was randomly mixed with synthetic data in various proportions. A classification experiment was conducted for each mix and the AUC was calculated.

In "out-of-time" experiments, the stability and improvement of AUC were investigated depending on the number of outliers generated from a Gaussian distribution at $3\sigma$ level in the macro-variables of the datasets. In these experiments, each dataset was split into training and testing sets based on time logs. The portion of the dataset prior to 2022 served as the training set, while data from and including 2022 formed the test set. The zGAN model was generated on the train set, after which 80 datasets were generated for each percentage of outliers in the macro-variables, including the option without outliers.

In the binary classification task, a classification model was trained for each set of synthetic data. The AUC was evaluated on a random sample of the test set that constituted 80\% of the entire test dataset. The final task involving 81 iterations of AUC computation for classification model training used a dataset composed of 80 datasets for each percentage of outliers, with the AUC being computed on the entire test set.

To assess the impact of outliers on AUC, median AUC, range, and interquartile range (IQR) were used as evaluation metrics.

\subsection{Correlation Analysis} To assess the ability of GANs to preserve and reproduce column correlations, the Pearson correlation coefficient was applied. Both the original dataset and the synthetic dataset were preprocessed as follows: missing values in numerical columns were filled with consistent values that significantly differed from the column’s mean and did not appear elsewhere in the data. In categorical columns, missing values were grouped into a separate category and all entries in categorical columns were encoded using LabelEncoder. Datetime columns were converted to Unix-time format. After these transformations, the dataset was normalized using StandardScaler.

To visually observe differences in the ability of various GANs to reproduce correlations, heatmaps of the differences between correlation matrices of the original and each of the synthetic datasets were used. A single-color scale was set for all heatmaps.

\subsection{Data Description}

The experiments utilized both private datasets with accessibility permissions authorized by zypl.ai and the open Titanic\footnote{https://www.kaggle.com/competitions/titanic/data} dataset. In the interest of original data owners, characteristics of the private datasets including names are concealed and designated by numbered Latin letters: $A_1$, ..., $A_9$. These real datasets constitute fully anonymized historical loan performance data across a variety of markets where zypl.ai operates and exhibit class imbalance in the target feature. The coefficient $k_{im}$ used to describe class imbalance represents the ratio of samples in the minority class in relation to the majority class.

Factors affecting the quality of synthetic data generation, in addition to correlations between dataset features, include the total number of features and the number of categorical features in the datasets.

The ratio of minority to majority classes, $k_{im}$, the total number of columns, the number of categorical columns and the availability of time-series column in private datasets are shown in Table~\ref{tab:private_datasets}.

\begin{table}[!htbp]
\centering
\renewcommand{\arraystretch}{2}
\begin{tabular}{|c|c|c|c|c|c|c|c|c|c|}
\hline
\textbf{Characteristics} & $\mathbf{A_1}$ & $\mathbf{A_2}$ & $\mathbf{A_3}$ & $\mathbf{A_4}$ & $\mathbf{A_5}$ & $\mathbf{A_6}$ & $\mathbf{A_7}$ & $\mathbf{A_8}$ & $\mathbf{A_9}$ \\
\hline
$k_{im}$ & 0.07 & 0.19 & 0.02 & 0.08 & 0.01 & 0.14 & 0.21 & 0.23 & 0.06 \\
\hline
Number of columns & 21 & 29 & 33 & 18 & 29 & 15 & 30 & 68 & 25 \\
\hline
Number of rows & 347K & 62K & 1.76M & 34K & 759K & 18K & 3189 & 139K & 2721 \\
\hline
Categorical columns & 6 & 10 & 16 & 18 & 13 & 6 & 15 & 10 & 9 \\
\hline
Number of categories & 56 & 44 & 370 & 60 & 44 & 29 & 242 & 47 & 67 \\
\hline
Time-series column & $\checkmark$ & $\checkmark$ & $\checkmark$ & $\times$ & $\times$ & $\times$ & $\checkmark$ & $\checkmark$ & $\checkmark$ \\
\hline
\end{tabular}
\vspace{0.3cm}
\caption{Characteristics of private datasets.}
\label{tab:private_datasets}
\end{table}

Experiments were also conducted on the open-source Titanic dataset. The training set of original Titanic dataset was split into train and test sets with a ratio of 0.33 consisting of 596 training records and 295 test records.

With ’Survived’ as the target variable, the following features were selected for further experiments: "Pclass", "Sex", "Age", "SibSp", "Parch", "Fare", "Cabin", "Embarked" with $k_{im}$ ratio of 0.62. The dataset contains a total of 9 features, 7 of which are categorical. In total, there are 173 categories in these columns.

For the training of all GAN models, the resulting training dataset, after splitting, was augmented by random sampling to a total of 10728 records. For other types of models, data augmentation was not applied, as these models were invariant to the number of training data.

\subsection{Model Definitions} The experiments included two GAN model variants (CTGAN and CopulaGAN) along with four alternative models (TVAE\footnote{https://docs.sdv.dev/sdv/single-table-data/modeling/synthesizers}, PrivBayes \footnote{https://github.com/hazy/dpart}, Synthpop\footnote{https://www.synthpop.org.uk/about-synthpop.html}, and GaussianCopula). CTGAN uses conditional probability to model complex dependencies in tabular data, making it a benchmark for tasks involving both numerical and categorical variables. TVAE offers a probabilistic approach based on autoencoders, allowing the modeling of complex distributions. These models provide strong neural network baselines for evaluating the performance of zGAN.

Classical and hybrid models such as Synthpop and PrivBayes that focus on data privacy, as well as GaussianCopula and CopulaGAN that utilize copula functions to model dependencies between variables, were incorporated in the comparative study. While GaussianCopula represents a traditional statistical method, CopulaGAN combines copula-based statistical modeling with neural network approaches. The diversity in selection of test models enables a comprehensive evaluation of how zGAN handles dependency modeling and synthetic data generation relative to existing methods.

On a dataset with simple correlation between features such as the Titanic dataset, the hyperparameters of the models available in the SDV library and Synthpop were set to their default values. 

In an experiment with the PrivBayes model on the Titanic dataset, the epsilon parameter which characterizes data privacy was set to 0.01. The PrivBayes model differs from other generative models used in the experiments by lacking built-in support for missing data. For experiments with this model, the data were preprocessed by imputing missing numerical features with consistent outlier negative values. Missing categorical features were treated as a separate category and subsequently encoded through Label Encoding. After imputing missing values and encoding categories, the data were normalized and the generated data were then reverse-transformed.

The Synthpop model’s inability to handle generating features with many categories, such as in the Titanic dataset where the Cabin feature has 109 categories, led to multi-category features getting encoded using one-hot encoding.

\section{Experimental Results}
\label{sec:experimental_results}

\subsection{Classification task}

\paragraph{Out-of-Sample experiments}

Table~\ref{tab:gan_comparison} presents the median AUC values obtained from experimental studies.

\begin{table}[!htbp]
\centering
\renewcommand{\arraystretch}{1.1}
\begin{tabular}{|c|c|c|c|c|c|}
\hline
\textbf{Dataset} & \textbf{zGAN} & \textbf{CTGAN} & \textbf{TVAE} & \textbf{CopulaGAN} & \textbf{Baseline} \\
\hline
$A_1$  & \textcolor{darkgreen}{0.7553} & \textcolor{red}{0.7496} & \textcolor{blue}{0.7504} & 0.6681 & 0.7841 \\
\hline
$A_2$  & \textcolor{darkgreen}{0.7972} & \textcolor{red}{0.7633} & 0.7428 & \textcolor{blue}{0.7959} & 0.8216 \\
\hline
$A_3$  & \textcolor{darkgreen}{0.7230} & \textcolor{blue}{0.7035} & \textcolor{red}{0.6461} & 0.6219 & 0.7552 \\
\hline
$A_4$  & \textcolor{darkgreen}{0.7095} & \textcolor{red}{0.6350} & 0.6244 & \textcolor{blue}{0.6663} & 0.7343 \\
\hline
$A_5$  & \textcolor{darkgreen}{0.8311} & \textcolor{blue}{0.8284} & 0.6163 & \textcolor{red}{0.6763} & 0.8680 \\
\hline
$A_6$ & \textcolor{darkgreen}{0.8120} & \textcolor{blue}{0.8025} & 0.7898 & \textcolor{red}{0.7942} & 0.8154 \\
\hline
$A_7$ & \textcolor{darkgreen}{0.7704} & 0.7013 & \textcolor{blue}{0.7279} & \textcolor{red}{0.7093} & 0.8183 \\
\hline
$A_8$ & \textcolor{blue}{0.7118} & 0.6992 & \textcolor{darkgreen}{0.7257} & \textcolor{red}{0.7010} & 0.7907 \\
\hline
$A_9$ & \textcolor{blue}{0.6525} & 0.5705 & \textcolor{darkgreen}{0.6874} & \textcolor{red}{0.5955} & 0.7109 \\
\hline
Titanic & \textcolor{darkgreen}{0.8163} & \textcolor{red}{0.7923} & 0.7874 & \textcolor{blue}{0.8154} & 0.8736 \\
\hline
\end{tabular}
\vspace{0.3cm}
\caption{The median AUC values are color-coded with in the following descending order among non-baseline models tested – \textcolor{darkgreen}{Green} indicates the highest value, \textcolor{blue}{Blue} the second highest, and \textcolor{red}{Red} the third highest in each row.}
\label{tab:gan_comparison}
\end{table}

As the results demonstrate, in 8 out of 10 backtests, zGAN outperforms every other model in terms of predictive accuracy except the baseline model trained on real data. On average, the AUC of zGAN is 0.03 points higher than that of CTGAN and 0.05 points higher than that of TVAE or CopulaGAN.

Extended results from these experiments are shown in Table~\ref{tab:appendix_median_auc}. With the main values representing median AUC, the values in parentheses stand for range of AUC variation as a difference between the minimum and maximum AUC values. If an experiment was not conducted, the cell is left empty. If an experiment was conducted but the generated synthetic data did not allow for training of a classification model, the cell is marked as "Undefined".

\paragraph{Out-of-Time experiments} OOT results of zGAN experiments involving proportional mixture of real and synthetic data are presented below in Table~\ref{tab:zGAN_auc_scores50}.

\begin{table}[!htbp]
\centering
\begin{tabular}{|l|c|c|c|c|c|c|}
\hline
\textbf{Datasets} & \textbf{100\% Synthetic} & \textbf{1:1} & \textbf{0.1:1} & \textbf{0.01:1} & \textbf{0.001:1} & \textbf{100\% Real Data} \\ \hline
$A_1$ & \makecell{\textcolor{darkgreen}{0.6738} \\ \scriptsize(0.6641 : 0.6812)} & \makecell{\textcolor{blue}{0.6684} \\ \scriptsize(0.6335 : 0.6862)} & \makecell{\textcolor{red}{0.6262} \\ \scriptsize(0.6017 : 0.6415)} & \makecell{0.6165 \\ \scriptsize(0.5979 : 0.6320)} & \makecell{0.6138 \\ \scriptsize(0.6014 : 0.6305)} & \makecell{0.6100 \\ \scriptsize(0.5996 : 0.6184)} \\ \hline
$A_2$ & \makecell{0.7862 \\ \scriptsize(0.7786 : 0.7917)} & \makecell{0.8108 \\ \scriptsize(0.8015 : 0.8156)} & \makecell{0.8157 \\ \scriptsize(0.8087 : 0.8204)} & \makecell{\textcolor{red}{0.8191} \\ \scriptsize(0.8138 : 0.8225)} & \makecell{\textcolor{blue}{0.8227} \\ \scriptsize(0.8178 : 0.8264)} & \makecell{\textcolor{darkgreen}{0.8238} \\ \scriptsize(0.8195 : 0.8279)} \\ \hline
$A_3$ & \makecell{0.7146 \\ \scriptsize(0.7101 : 0.7186)} & \makecell{0.7302 \\ \scriptsize(0.7264 : 0.7339)} & \makecell{0.7433 \\ \scriptsize(0.7389 : 0.7472)} & \makecell{\textcolor{red}{0.7435} \\ \scriptsize(0.7395 : 0.7462)} & \makecell{\textcolor{blue}{0.7444} \\ \scriptsize(0.7393 : 0.7479)} & \makecell{\textcolor{darkgreen}{0.7456} \\ \scriptsize(0.7417 : 0.7488)} \\ \hline
$A_7$ & \makecell{\textcolor{darkgreen}{0.8250} \\ \scriptsize(0.8134 : 0.8517)} & \makecell{\textcolor{blue}{0.8219} \\ \scriptsize(0.8050 : 0.8446)} & \makecell{\textcolor{red}{0.8092} \\ \scriptsize(0.7854 : 0.8366)} & \makecell{0.8091 \\ \scriptsize(0.7812 : 0.8295)} & \makecell{0.8076 \\ \scriptsize(0.7830 : 0.8241)} & \makecell{0.8091 \\ \scriptsize(0.7847 : 0.8378)} \\ \hline
$A_8$ & \makecell{0.7686 \\ \scriptsize(0.7628 : 0.7725)} & \makecell{0.7804 \\ \scriptsize(0.7767 : 0.7834)} & \makecell{0.7853 \\ \scriptsize(0.7787 : 0.7893)} & \makecell{\textcolor{blue}{0.7859} \\ \scriptsize(0.7807 : 0.7900)} & \makecell{\textcolor{darkgreen}{0.7865} \\ \scriptsize(0.7824 : 0.7903)} & \makecell{\textcolor{blue}{0.7859} \\ \scriptsize(0.7799 : 0.7895)} \\ \hline
$A_9$ & \makecell{\textcolor{blue}{0.7117} \\ \scriptsize(0.6674 : 0.7747)} & \makecell{\textcolor{darkgreen}{0.7233} \\ \scriptsize(0.6660 : 0.7805)} & \makecell{0.6443 \\ \scriptsize(0.5628 : 0.7069)} & \makecell{0.6627 \\ \scriptsize(0.5789 : 0.7597)} & \makecell{\textcolor{red}{0.7046} \\ \scriptsize(0.6568 : 0.7567)} & \makecell{0.6862 \\ \scriptsize(0.6129 : 0.7429)} \\ \hline
\end{tabular}
\vspace{0.3cm}
\caption{AUC out-of-time validation on 50\% dataset of zGAN – \textcolor{darkgreen}{Green} indicates the highest value, \textcolor{blue}{Blue} the second highest, and \textcolor{red}{Red} the third highest in each row.}
\label{tab:zGAN_auc_scores50}
\end{table}

Experimental results demonstrate that in 4 out of 6 iterations, models trained either solely on zGAN-generated data or containing a mixture of zGAN-generated and real data produce optimal AUC performance. In cases of $A_1$, $A_7$, and $A_9$ datasets, the median AUC value increases significantly when using zGAN-generated data or when adding a proportionally large amount of synthetic data to the real data. In the case of $A_1$, AUC increases by up to 0.064 points.

Table~\ref{tab:auc_scores80} shows the results of similar experiments with the dataset split into 80\% training. In most cases, the results show an increase in the median AUC value when synthetic data is added to the training set. In cases of $A_3$ and $A_9$ datasets, AUC decreases when mixing the data, with $A_3$ showing an insignificant decrease of 0.0002 points. In the remaining cases, AUC increases – for instance, in case of $A_1$, AUC increases significantly by 0.0189 points.

Comparative results of OOT experiments derived from CTGAN-generated datasets are shown in Table~\ref{tab:ctGAN_auc_scores50} below.

\begin{table}[!htbp]
    \centering
    \begin{tabular}{|l|c|c|c|c|c|c|}
        \hline
        \textbf{Datasets} & \textbf{100\% Synthetic} & \textbf{1:1} & \textbf{0.1:1} & \textbf{0.01:1} & \textbf{0.001:1} & \textbf{100\% Real Data} \\
        \hline
        $A_1$ & \makecell{\textcolor{darkgreen}{0.7077} \\ \scriptsize(0.6959 : 0.7151)} & \makecell{\textcolor{blue}{0.7054} \\ \scriptsize(0.6675 : 0.7144)} & \makecell{\textcolor{red}{0.6295} \\ \scriptsize(0.6137 : 0.6678)} & \makecell{0.6177 \\ \scriptsize(0.6047 : 0.6338)} & \makecell{0.6103 \\ \scriptsize(0.6018 : 0.6196)} & \makecell{0.6100 \\ \scriptsize(0.5996 : 0.6184)} \\
        \hline
        $A_2$ & \makecell{0.7815 \\ \scriptsize(0.5000 : 0.7933)} & \makecell{0.8063 \\ \scriptsize(0.8016 : 0.8111)} & \makecell{0.8199 \\ \scriptsize(0.8140 : 0.8245)} & \makecell{\textcolor{red}{0.8235} \\ \scriptsize(0.8184 : 0.8275)} & \makecell{\textcolor{blue}{0.8237} \\ \scriptsize(0.8189 : 0.8287)} & \makecell{\textcolor{darkgreen}{0.8238} \\ \scriptsize(0.8195 : 0.8279)} \\
        \hline
        $A_3$ & \makecell{0.7025 \\ \scriptsize(0.6884 : 0.7109)} & \makecell{0.7368 \\ \scriptsize(0.7317 : 0.7396)} & \makecell{\textcolor{red}{0.7452} \\ \scriptsize(0.7425 : 0.7478)} & \makecell{\textcolor{blue}{0.7453} \\ \scriptsize(0.7414 : 0.7488)} & \makecell{\textcolor{darkgreen}{0.7456} \\ \scriptsize(0.7427 : 0.7491)} & \makecell{\textcolor{darkgreen}{0.7456} \\ \scriptsize(0.7417 : 0.7488)} \\
        \hline
        $A_7$ & \makecell{0.7707 \\ \scriptsize(0.7456 : 0.7961)} & \makecell{\textcolor{darkgreen}{0.8117} \\ \scriptsize(0.7951 : 0.8370)} & \makecell{\textcolor{blue}{0.8108} \\ \scriptsize(0.7862 : 0.8406)} & \makecell{0.8090 \\ \scriptsize(0.7830 : 0.8324)} & \makecell{0.8079 \\ \scriptsize(0.7802 : 0.8363)} & \makecell{\textcolor{red}{0.8091} \\ \scriptsize(0.7847 : 0.8378)} \\
        \hline
        $A_8$ & \makecell{0.7481 \\ \scriptsize(0.7409 : 0.7527)} & \makecell{0.7761 \\ \scriptsize(0.7700 : 0.7812)} & \makecell{0.7814 \\ \scriptsize(0.7772 : 0.7857)} & \makecell{\textcolor{red}{0.7855} \\ \scriptsize(0.7798 : 0.7892)} & \makecell{\textcolor{darkgreen}{0.7862} \\ \scriptsize(0.7822 : 0.7898)} & \makecell{\textcolor{blue}{0.7859} \\ \scriptsize(0.7799 : 0.7895)} \\
        \hline
        $A_9$ & \makecell{0.6861 \\ \scriptsize(0.6731 : 0.7050)} & \makecell{0.7072 \\ \scriptsize(0.6725 : 0.7427)} & \makecell{\textcolor{darkgreen}{0.7136} \\ \scriptsize(0.7039 : 0.7214)} & \makecell{\textcolor{red}{0.7094} \\ \scriptsize(0.6897 : 0.7202)} & \makecell{\textcolor{blue}{0.7104} \\ \scriptsize(0.7014 : 0.7283)} & \makecell{0.6862 \\ \scriptsize(0.6129 : 0.7429)} \\
        \hline
    \end{tabular}
    \vspace{0.3cm}
    \caption{AUC out-of-time validation on 50\% dataset of CTGAN – \textcolor{darkgreen}{Green} indicates the highest value, \textcolor{blue}{Blue} the second highest, and \textcolor{red}{Red} the third highest in each row.}
    \label{tab:ctGAN_auc_scores50}
\end{table}

In the OOT experiments on mixing real and synthetic data using CTGAN, the effect of mixing on the median AUC was demonstrated. In the case of $A_2$ dataset, the AUC calculated on real data slightly exceeded the highest AUC obtained through data mixing by 0.001 points, while in the case of $A_3$ no AUC improvement was observed through mixing. In the cases of $A_1$, $A_7$, $A_8$, and $A_9$ datasets, there was an increase in AUC with mixing – with the largest increase of 0.0977 observed in the case of $A_1$ dataset.

The OOT experiments on mixing real training and synthetic data show, in most cases, an increase in AUC, which also indicates improved stability in the classification task. The increase in median AUC on synthetic data or when mixing synthetic data with real data is observed in experiments involving both CTGAN and zGAN. This is reflective of the hypothesis that the improvement in classification model performance occurs invariantly of the GAN used for generating synthetic data.

The results of OOT experiments with synthetic outliers as part of macro data of synthetic datasets generated by zGAN are shown in Table~\ref{tab:auc_scores_outliers}, reflecting the case of dataset $A_9$ that characterizes a rapidly growing economy.

\begin{table}[!htbp]
\centering
\renewcommand{\arraystretch}{1.2}
\begin{tabular}{|c|c|}
\hline
\textbf{Percent of outliers} & \textbf{Median AUC (Range)} \\ \hline
100\%   & 0.6862 (0.5706:0.7947)  \\ \hline
50\%    & 0.6867 (0.5669:0.7677) \\ \hline
10\%    & 0.6962 (0.5736:0.7769) \\ \hline
7.7\%   & 0.7061 (0.6357:0.7819) \\ \hline
\textcolor{blue}{7.4\%}   & \textcolor{blue}{0.7122} (0.5985:0.8048) \\ \hline
7.1\%   & 0.7077 (0.6039:0.7770) \\ \hline
7\%     & 0.7095 (0.6170:0.7948) \\ \hline
6.9\%   & 0.7074 (0.6163:0.7807) \\ \hline
6.6\%   & 0.7079 (0.5985:0.7650) \\ \hline
6.3\%   & 0.6966 (0.5838:0.7875) \\ \hline
6\%     & 0.7096 (0.6371:0.7853) \\ \hline
\textcolor{darkgreen}{5\%}     & \textcolor{darkgreen}{0.7147} (0.5858:0.7890) \\ \hline
\textcolor{red}{3\%}     & \textcolor{red}{0.7116} (0.6130:0.7694) \\ \hline
1\%     & 0.7032 (0.6027:0.7736) \\ \hline
\textbf{Without} & \textbf{0.7020} (0.6154:0.8194) \\ \hline
\end{tabular}
\vspace{0.3cm}
\caption{Median AUC values for different percentages of outliers in macro-features for $A_9$ - \textcolor{darkgreen}{Green} indicates the highest value, \textcolor{blue}{Blue} the second highest, and \textcolor{red}{Red} the third highest.}
\label{tab:auc_scores_outliers}
\end{table}

Experiments with synthetic outliers show an increase in the median AUC in most cases, without a clear linear dependence on the percentage of outliers in the macro data, as shown in Figure~\ref{fig:median_auc_outliers}. When the number of synthetic outliers exceeds 10\% of the total number of records in the macro data, the median AUC decreases. The largest increase in median AUC is observed with 3\%, 5\%, and 7\% outliers in the macro data columns.

The change in median AUC when adding synthetic outliers relative to the case without outliers, as well as the results of the Wilcoxon test for the significance of AUC changes, are shown in Table~\ref{tab:auc_significant_test}. A significant decrease in AUC is observed when the number of outliers in the synthetic macro data reaches 50\% or more. The maximum and statistically significant increase in AUC is observed with the threshold of 5\% outliers generated in the synthetic macro data.

Experimental results with generated synthetic outliers demonstrate the usefulness of outliers for improving the quality of classification task results along with stability of zGAN in generating synthetic data with synthetic outliers.

\subsection{Correlation Analysis}

Figure~\ref{fig:selected_difference_heatmaps} below shows the difference in Pearson correlation matrices between synthetic datasets generated by various GANs and the original Titanic dataset. The correlation matrices and differences in correlations for other GANs are shown in Figures~\ref{fig:heatmaps} and ~\ref{fig:difference_heatmaps}, respectively.

\begin{figure}[!htbp]
    \centering
    \resizebox{0.5\linewidth}{!}{\includegraphics{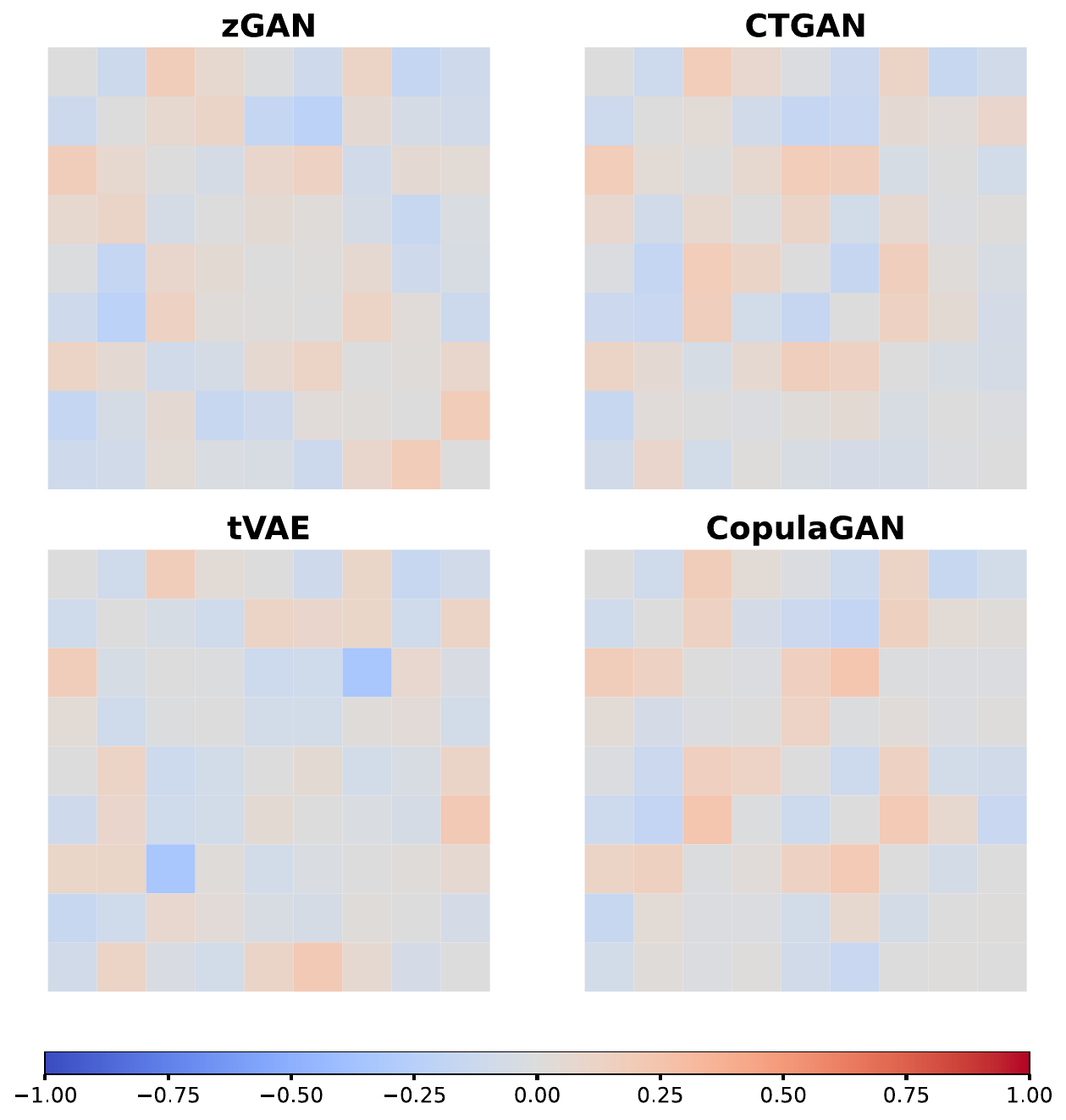}}
    \caption{Correlation difference heatmaps between synthetic and original Titanic datasets.}
    \label{fig:selected_difference_heatmaps}
\end{figure}

The heatmaps provided in Figure~\ref{fig:selected_difference_heatmaps} illustrate differences between the correlation matrices of synthetic datasets generated by zGAN, CTGAN, TVAE, and CopulaGAN in comparison to the original dataset. TVAE shows moderate differences, indicating adequate preservation of the original correlations. CopulaGAN exhibits more significant variations, suggesting a less reliable retention of the original structure. CTGAN and zGAN demonstrate the smallest differences, with zGAN surpassing CTGAN – indicating the highest fidelity in maintaining the original correlation structure. Therefore, CTGAN and zGAN are found to be the most effective in generating synthetic data that closely mirrors the original dataset followed by TVAE, which offers an adequate balance between fidelity and quality in preserving correlations, and CopulaGAN as the least effective.

\section*{Conclusions}
\label{sec:conclusions}

While synthetic data generation has been traditionally utilized to resolve data insufficiency and/or privacy challenges, applications of synthetic data to derive model performance uplifts have been scarce. This experimental study has explored both the relative realism and uplift capabilities of zGAN, a novel model architecture devised to generate synthetic tabular data with a focus on outliers. 

The similarity filter applied in zGAN ensures that no real data leaks into synthetic datasets. The embedded capability of zGAN to generate outliers enables the modeling of highly uncertain economic events or the augmentation of outliers for further training of predictive models. This can be additionally applied for tasks involving the processing, detection or removal of outliers. The zGAN model can generate outliers automatically based on the covariance of features in real data.

Out-of-sample experiments comparing the classification performance of synthetic data generated by state-of-the-art open-source GANs and classical generative models demonstrate the advantage of zGAN over the evaluated models. On average, the AUC of zGAN surpasses that of CTGAN by 0.03 points, TVAE by 0.05 points, and CopulaGAN by 0.06 points.

The application of zGAN for generating synthetic tabular data allows the classification model to learn underlying patterns without interacting with real client data. Out-of-time experiments demonstrate that training of a classification model on synthetic data improves the model’s generalization ability, positively impacting the quality of performance when dealing with future, previously unseen data. In an experiment to enrich and mix real data with synthetic data for $A_1$ sample, an increase in median AUC of 0.0638 points was achieved when using only synthetic data for training. Furthermore, an increase of 0.0584 points of AUC was obtained when mixing training of real data with synthetic data in a 1:1 ratio.

Generation of synthetic outliers in the OOT case of $A_9$ sample, where the pre-2022 data was used for training and the 2022-2023 period data was used for testing, leads to a significant improvement in classification performance. In the experiment on generating synthetic outliers in the macro-variables of synthetic dataset, an increase in median AUC was achieved with a moderate volume of synthetic outlier generation of up to 10\%. The most significant AUC improvement of 0.0127 points was observed at 5\% outlier generation threshold. These results highlight importance of precise outliers balancing in out-of-time predictions where outliers tend to have strong predictive impact.

A comparison of correlations between generated and real datasets using the open Titanic sample demonstrates the ability of generative models to reproduce correlations. All GAN models and TVAE show the most accurate reproduction of column correlations, while models based on classical machine learning principles demonstrate the poorest performance. The patterns of column correlation differences are similar for GaussianCopula and PrivBayes, with the former showing smaller deviations. TVAE and SynthPop exhibit unique patterns compared to other models and each other, with TVAE performing significantly better and closely matching the quality of GAN models. The most similar and high-quality deviation patterns are observed in CTGAN, CopulaGAN, and zGAN.

Future experimental work can be conducted on testing the relative performance of zGAN vis-à-vis other proprietary or open-source models including novel GANs. Furthermore, exploration of zGAN’s applicability can be conducted in additional fields of interest and on model targets beyond classification. The study of zGAN’s prospective multimodality beyond tabular data can scale the conceptual scope of model’s usefulness in other areas of artificial intelligence. 

\paragraph{Author Contributions:} Conceptualization, Azimi; zGAN development, Khalilbekov, Nizamitdinov and Shulgin; methodology, Khalilbekov and Shulgin; data curation, Khalilbekov; formal analysis, Varshavskiy and Shulgin; experi- ment implementation, Varshavskiy and Shulgin; validation, Khalilbekov and Nizamitdinov; visualization, Varshavskiy and Shulgin; writing—original draft, Varshavskiy and Shulgin; writing—review and editing, Azimi, Khalilbekov and Nizamitdinov; project management, Boboeva and Khalilbekov; project administration, Azimi and Noyoftova. All authors have read and agreed to the published version of the manuscript. 

\paragraph{Acknowledgements:} This research study was conducted under the initiative of zypl.ai\footnote{https://zypl.ai/en} and its corresponding staff, enterprise partners and investors. Development of zGAN has been publicly supported by the Artificial Intelligence Council\footnote{https://aic.tj/en} under the Ministry of Industry and New Technologies of the Republic of Tajikistan. Recognized as a foundational research \& development effort emerging from Tajikistan, the zGAN project serves to enhance implementation of the national "Strategy for Artificial Intelligence Development in the Republic of Tajikistan until 2040". Forthcoming exploration of intellectual property rights and commercial applications of zGAN have been reserved for zypl.ai (in the financial services domain) and zehn.ai\footnote{https://zehnlab.ai/en/about} (in non-financial services domains).

\paragraph{Conflicts of Interest:} The authors declare no conflict of interest.

\bibliographystyle{unsrt}
\bibliography{references}

\begin{thebibliography}{10}

\bibitem{Assefa2021}
Samuel~A. Assefa, Danial Dervovic, Mahmoud Mahfouz, Robert~E. Tillman, Prashant Reddy, and Manuela Veloso.
\newblock Generating synthetic data in finance: opportunities, challenges and pitfalls.
\newblock In {\em Proceedings of the First ACM International Conference on AI in Finance}, ICAIF '20, New York, NY, USA, 2021. Association for Computing Machinery.

\bibitem{Shokri2017}
R.~Shokri, M.~Stronati, C.~Song, and V.~Shmatikov.
\newblock Membership inference attacks against machine learning models.
\newblock In {\em 2017 IEEE Symposium on Security and Privacy (SP)}, pages 3--18, Los Alamitos, CA, USA, may 2017. IEEE Computer Society.

\bibitem{Bluwstein2023}
Kristina Bluwstein, Marcus Buckmann, Andreas Joseph, Sujit Kapadia, and Özgür Şimşek.
\newblock Credit growth, the yield curve and financial crisis prediction: Evidence from a machine learning approach.
\newblock {\em Journal of International Economics}, 145:103773, 2023.

\bibitem{Ghil2011}
M.~Ghil, P.~Yiou, S.~Hallegatte, B.~D. Malamud, P.~Naveau, A.~Soloviev, P.~Friederichs, V.~Keilis-Borok, D.~Kondrashov, V.~Kossobokov, O.~Mestre, C.~Nicolis, H.~W. Rust, P.~Shebalin, M.~Vrac, A.~Witt, and I.~Zaliapin.
\newblock Extreme events: dynamics, statistics and prediction.
\newblock {\em Nonlinear Processes in Geophysics}, 18(3):295--350, 2011.

\bibitem{Ajagbe2024}
Sunday~Adeola Ajagbe and Matthew~O. Adigun.
\newblock Deep learning techniques for detection and prediction of pandemic diseases: a systematic literature review.
\newblock {\em Multimedia Tools and Applications}, 83:5893--5927, 1 2024.

\bibitem{Wabartha2020}
Maxime Wabartha, Audrey Durand, Vincent François-Lavet, and Joelle Pineau.
\newblock Handling black swan events in deep learning with diversely extrapolated neural networks.
\newblock In Christian Bessiere, editor, {\em Proceedings of the Twenty-Ninth International Joint Conference on Artificial Intelligence, {IJCAI-20}}, pages 2140--2147. International Joint Conferences on Artificial Intelligence Organization, 7 2020.
\newblock Main track.

\bibitem{goodfellow2014}
Ian~J. Goodfellow, Jean Pouget-Abadie, Mehdi Mirza, Bing Xu, David Warde-Farley, Sherjil Ozair, Aaron Courville, and Yoshua Bengio.
\newblock Generative adversarial networks, 2014.

\bibitem{Bourou2021}
Stavroula Bourou, Andreas~El Saer, Terpsichori-Helen Velivassaki, Artemis Voulkidis, and Theodore Zahariadis.
\newblock A review of tabular data synthesis using gans on an ids dataset.
\newblock {\em Information}, 12:375, 9 2021.

\bibitem{Nayak2024}
Sasmita~Manjari Nayak and Minakhi Rout.
\newblock Impact of different outlier handling techniques on gan based hybrid bankruptcy prediction models.
\newblock {\em Indian Journal Of Science And Technology}, 17:373--385, 1 2024.

\bibitem{deHaan2006}
Laurens de~Haan and Ana Ferreira.
\newblock {\em Extreme Value Theory}.
\newblock Springer New York, 2006.

\bibitem{Zhang2018}
Han Zhang, Ian Goodfellow, Dimitris Metaxas, and Augustus Odena.
\newblock Self-attention generative adversarial networks.
\newblock In Kamalika Chaudhuri and Ruslan Salakhutdinov, editors, {\em Proceedings of the 36th International Conference on Machine Learning}, volume~97 of {\em Proceedings of Machine Learning Research}, pages 7354--7363. PMLR, 09--15 Jun 2019.

\bibitem{Yang2021}
Yu~Yang, Lei Sun, Xiuqing Mao, Leyu Dai, Song Guo, and Peiyuan Liu.
\newblock Using generative adversarial networks based on dual attention mechanism to generate face images.
\newblock In {\em 2021 International Conference on Computer Technology and Media Convergence Design (CTMCD)}, pages 14--19. IEEE, 4 2021.

\bibitem{Sun2021}
Guangcong Sun, Shifei Ding, Tongfeng Sun, and Chenglong Zhang.
\newblock Sa-capsgan: Using capsule networks with embedded self-attention for generative adversarial network.
\newblock {\em Neurocomputing}, 423:399--406, 1 2021.

\bibitem{Li2021}
Yushi Li and George Baciu.
\newblock Hsgan: Hierarchical graph learning for point cloud generation.
\newblock {\em IEEE Transactions on Image Processing}, 30:4540--4554, 2021.

\bibitem{taleb_black_2008}
Nassim~Nicholas Taleb.
\newblock {\em The Black Swan: The Impact of the Highly Improbable}.
\newblock Random House, London, 1 edition, 2008.

\bibitem{taleb2012antifragile}
N.N. Taleb.
\newblock {\em Antifragile: Things that Gain from Disorder}.
\newblock Penguin Books Limited, 2012.

\bibitem{Zhang2021}
Liang Zhang, Jianqing Wu, Jun Shen, Ming Chen, Rui Wang, Xinliang Zhou, Cankun Xu, Quankai Yao, and Qiang Wu.
\newblock Satp-gan: self-attention based generative adversarial network for traffic flow prediction.
\newblock {\em Transportmetrica B: Transport Dynamics}, 9:552--568, 1 2021.

\bibitem{Oh2021}
Eunkyu Oh, Taehun Kim, Yunhu Ji, and Sushil Khyalia.
\newblock Sting: Self-attention based time-series imputation networks using gan.
\newblock In {\em 2021 IEEE International Conference on Data Mining (ICDM)}. IEEE, December 2021.

\bibitem{Paul2021}
Paul Jeha, Michael Bohlke{-}Schneider, Pedro Mercado, Rajbir{-}Singh Nirwan, Shubham Kapoor, Valentin Flunkert, Jan Gasthaus, and Tim Januschowski.
\newblock {PSA-GAN:} progressive self attention gans for synthetic time series.
\newblock {\em CoRR}, abs/2108.00981, 2021.

\bibitem{Xu2018}
Lei Xu and Kalyan Veeramachaneni.
\newblock Synthesizing tabular data using generative adversarial networks, 2018.

\bibitem{Li2023_2}
Zijian Li and Zhihui Wang.
\newblock A self-attention-based differentially private tabular gan with high data utility, 2023.

\bibitem{Kossen2021}
Jannik Kossen, Neil Band, Clare Lyle, Aidan~N Gomez, Thomas Rainforth, and Yarin Gal.
\newblock Self-attention between datapoints: Going beyond individual input-output pairs in deep learning.
\newblock In M~Ranzato, A~Beygelzimer, Y~Dauphin, P~S Liang, and J~Wortman Vaughan, editors, {\em Advances in Neural Information Processing Systems}, volume~34, pages 28742--28756. Curran Associates, Inc., 2021.

\bibitem{Somepalli2021}
Gowthami Somepalli, Micah Goldblum, Avi Schwarzschild, C.~Bayan Bruss, and Tom Goldstein.
\newblock {SAINT:} improved neural networks for tabular data via row attention and contrastive pre-training.
\newblock {\em CoRR}, abs/2106.01342, 2021.

\bibitem{Mrini2019}
Khalil Mrini, Franck Dernoncourt, Trung Bui, Walter Chang, and Ndapa Nakashole.
\newblock Rethinking self-attention: An interpretable self-attentive encoder-decoder parser.
\newblock {\em CoRR}, abs/1911.03875, 2019.

\bibitem{Skrlj2020}
Blaz Skrlj, Saso Dzeroski, Nada Lavrac, and Matej Petkovic.
\newblock Feature importance estimation with self-attention networks.
\newblock {\em CoRR}, abs/2002.04464, 2020.

\bibitem{Sun2022}
Chang Sun, Johan van Soest, and Michel Dumontier.
\newblock Improving correlation capture in generating imbalanced data using differentially private conditional gans, 2022.

\bibitem{Vu2024}
Minh~H. Vu, Daniel Edler, Carl Wibom, Tommy Löfstedt, Beatrice Melin, and Martin Rosvall.
\newblock A correlation- and mean-aware loss function and benchmarking framework to improve gan-based tabular data synthesis, 2024.

\bibitem{Pavlov2024}
Arie Pavlov, Edita Grolman, Ikuya Morikawa, Toshiya Shimizu, Asaf Shabtai, and Yuval Elovici.
\newblock Tab2gan: Utilizing image conversion and gan inversion for tabular model robustness, 2024.

\bibitem{Zhao2022}
Zilong Zhao, Robert Birke, and Lydia~Y. Chen.
\newblock Fct-gan: Enhancing table synthesis via fourier transform, 2022.

\bibitem{Chattoraj2024}
Joyjit Chattoraj, Jian~Cheng Wong, Zhang Zexuan, Manna Dai, Xia Yingzhi, Li~Jichao, Xu~Xinxing, Ooi~Chin Chun, Yang Feng, Dao~My Ha, and Liu Yong.
\newblock Tailoring generative adversarial networks for smooth airfoil design, 2024.

\bibitem{Fallahian2024}
Mohammadali Fallahian, Mohsen Dorodchi, and Kyle Kreth.
\newblock Gan-based tabular data generator for constructing synopsis in approximate query processing: Challenges and solutions.
\newblock {\em Machine Learning and Knowledge Extraction}, 6:171--198, 1 2024.

\bibitem{Trindade2024}
Carolina Trindade, Luís Antunes, Tânia Carvalho, and Nuno Moniz.
\newblock Synthetic data outliers: Navigating identity disclosure, 2024.

\bibitem{Du2024}
Xusheng Du, Jiaying Chen, Jiong Yu, Shu Li, and Qiyin Tan.
\newblock Generative adversarial nets for unsupervised outlier detection.
\newblock {\em Expert Systems with Applications}, 236:121161, 2 2024.

\bibitem{Qi2024}
Sibo Qi, Juan Chen, Peng Chen, Peian Wen, Xianhua Niu, and Lei Xu.
\newblock An efficient gan-based predictive framework for multivariate time series anomaly prediction in cloud data centers.
\newblock {\em The Journal of Supercomputing}, 80:1268--1293, 1 2024.

\bibitem{Fang2024}
Yetong Fang.
\newblock Apib-gan: A generative adversarial networks based approach for anomaly prediction of internet behavior.
\newblock {\em Physical Communication}, 64:102315, 6 2024.

\bibitem{Oh2019}
Joo-Hyuk Oh, Jae~Yeol Hong, and Jun-Geol Baek.
\newblock Oversampling method using outlier detectable generative adversarial network.
\newblock {\em Expert Systems with Applications}, 133:1--8, 11 2019.

\bibitem{Aftabi2023}
Seyyede~Zahra Aftabi, Ali Ahmadi, and Saeed Farzi.
\newblock Fraud detection in financial statements using data mining and gan models.
\newblock {\em Expert Systems with Applications}, 227:120144, 10 2023.

\bibitem{Zhao2024}
Penghui Zhao, Zhongjun Ding, Yang Li, Xiaohan Zhang, Yuanqi Zhao, Hongjun Wang, and Yang Yang.
\newblock Sgad-gan: Simultaneous generation and anomaly detection for time-series sensor data with generative adversarial networks.
\newblock {\em Mechanical Systems and Signal Processing}, 210:111141, 3 2024.

\bibitem{pourreza2020}
Masoud Pourreza, Bahram Mohammadi, Mostafa Khaki, Samir Bouindour, Hichem Snoussi, and Mohammad Sabokrou.
\newblock G2d: Generate to detect anomaly, 2020.

\bibitem{Zakharov2024}
Kirill Zakharov, Elizaveta Stavinova, and Anton Lysenko.
\newblock Trgan: A time-dependent generative adversarial network for synthetic transactional data generation.
\newblock In {\em Proceedings of the 2023 7th International Conference on Software and E-Business}, ICSeB '23, page 1–8, New York, NY, USA, 2024. Association for Computing Machinery.

\end{thebibliography}

\appendix

\renewcommand{\thefigure}{\thesection.\arabic{figure}}
\renewcommand{\thetable}{\thesection.\arabic{table}}

\newpage
\section{Appendix: Classification task}
\label{sec:appendix_classification task}

\setcounter{figure}{0}
\setcounter{table}{0}

\begin{table}[!htbp]
\centering
\rotatebox{90}{
\begin{tabular}{|c|c|c|c|c|c|c|c|c|c|}
\hline
\textbf{Datasets} & \textbf{zGAN} & \textbf{CTGAN} & \textbf{TVAE} & \textbf{Synthpop} & \textbf{PrivBayes} & \makecell{\textbf{Gaussian}\\\textbf{Copula}} & \textbf{CopulaGAN} & \textbf{Baseline} \\
\hline
$A_1$ & \makecell{0.7553 \\ \scriptsize(0.7496 : 0.7605)} & \makecell{0.7496 \\ \scriptsize(0.7437 : 0.7545)} & \makecell{0.7504 \\ \scriptsize(0.7460 : 0.7548)} & \makecell{0.7625 \\ \scriptsize(0.7584 : 0.7664)} & \makecell{0.5123 \\ \scriptsize(0.4645 : 0.6001)} & \makecell{0.5747 \\ \scriptsize(0.5695 : 0.5797)} & \makecell{0.6681 \\ \scriptsize(0.6629 : 0.6739)} & \makecell{0.7841 \\ \scriptsize(0.7764 : 0.7884)} \\
\hline
$A_2$ & \makecell{0.7972 \\ \scriptsize(0.7920 : 0.8026)} & \makecell{0.7633 \\ \scriptsize(0.7551 : 0.7777)} & \makecell{0.7428 \\ \scriptsize(0.7359 : 0.7627)} & \makecell{0.8127 \\ \scriptsize(0.8084 : 0.8181)} & \makecell{0.5511 \\ \scriptsize(0.4365 : 0.6461)} & \makecell{0.7743 \\ \scriptsize(0.7678 : 0.7799)} & \makecell{0.7959 \\ \scriptsize(0.7919 : 0.8005)} & \makecell{0.8216 \\ \scriptsize(0.8176 : 0.8265)} \\
\hline
$A_3$ & \makecell{0.7230 \\ \scriptsize(0.7191 : 0.7272)} & \makecell{0.7035 \\ \scriptsize(0.6994 : 0.7071)} & \makecell{0.6461 \\ \scriptsize(0.6314 : 0.6627)} & Undefined & Undefined & \makecell{0.5101 \\ \scriptsize(0.5067 : 0.5135)} & \makecell{0.6219 \\ \scriptsize(0.6173 : 0.6252)} & \makecell{0.7552 \\ \scriptsize(0.7471 : 0.7587)} \\
\hline
$A_4$ & \makecell{0.7095 \\ \scriptsize(0.6946 : 0.7233)} & \makecell{0.6350 \\ \scriptsize(0.6216 : 0.6494)} & \makecell{0.6244 \\ \scriptsize(0.5926 : 0.6466)} & \makecell{0.6567 \\ \scriptsize(0.6414 : 0.6685)} & \makecell{0.5664 \\ \scriptsize(0.5200 : 0.6081)} & \makecell{0.5650 \\ \scriptsize(0.5473 : 0.5837)} & \makecell{0.6663 \\ \scriptsize(0.6513 : 0.6828)} & \makecell{0.7343 \\ \scriptsize(0.7135 : 0.7470)} \\
\hline
$A_5$ & \makecell{0.8311 \\ \scriptsize(0.8216 : 0.8392)} & \makecell{0.8284 \\ \scriptsize(0.8216 : 0.8388)} & \makecell{0.6163 \\ \scriptsize(0.5880 : 0.6505)} & \makecell{0.7402 \\ \scriptsize(0.7373 : 0.7437)} & Undefined & \makecell{0.5232 \\ \scriptsize(0.5185 : 0.5264)} & \makecell{0.6763 \\ \scriptsize(0.6732 : 0.6830)} & \makecell{0.8680 \\ \scriptsize(0.8603 : 0.8725)} \\
\hline
$A_6$ & \makecell{0.8120 \\ \scriptsize(0.8008 : 0.8310)} & \makecell{0.8025 \\ \scriptsize(0.7913 : 0.8204)} & \makecell{0.7898 \\ \scriptsize(0.7642 : 0.8126)} & \makecell{0.7911 \\ \scriptsize(0.7809 : 0.8043)} & \makecell{0.5570 \\ \scriptsize(0.4410 : 0.6852)} & \makecell{0.7221 \\ \scriptsize(0.7084 : 0.7393)} & \makecell{0.7942 \\ \scriptsize(0.7828 : 0.8061)} & \makecell{0.8154 \\ \scriptsize(0.8039 : 0.8325)} \\
\hline
$A_7$ & \makecell{0.7704 \\ \scriptsize(0.7453 : 0.7930)} & \makecell{0.7013 \\ \scriptsize(0.6769 : 0.7239)} & \makecell{0.7279 \\ \scriptsize(0.6951 : 0.7558)} & \makecell{0.7804 \\ \scriptsize(0.7596 : 0.8038)} & \makecell{0.5470 \\ \scriptsize(0.5011 : 0.6162)} & \makecell{0.7256 \\ \scriptsize(0.6923 : 0.7520)} & \makecell{0.7093 \\ \scriptsize(0.6654 : 0.7403)} & \makecell{0.8183 \\ \scriptsize(0.7946 : 0.8419)} \\
\hline
$A_8$ & \makecell{0.7118 \\ \scriptsize(0.7044 : 0.7204)} & \makecell{0.6992 \\ \scriptsize(0.6903 : 0.7113)} & \makecell{0.7257 \\ \scriptsize(0.7156 : 0.7359)} & Undefined & \makecell{0.5747 \\ \scriptsize(0.4924 :  0.6609)} & \makecell{0.6836 \\ \scriptsize(0.6779 :  0.6886)} & \makecell{0.7010 \\ \scriptsize(0.7067 : 0.7136)} & \makecell{0.7907 \\ \scriptsize(0.7826 : 0.7971)} \\
\hline
$A_9$ & \makecell{0.6525 \\ \scriptsize(0.5826 : 0.7163)} & \makecell{0.5705 \\ \scriptsize(0.5056 : 0.6510)} & \makecell{0.6874 \\ \scriptsize(0.6091 : 0.7618)} & \makecell{0.6739 \\ \scriptsize(0.5999 : 0.7127)} & \makecell{0.5649 \\ \scriptsize(0.5283 : 0.6163)} & \makecell{0.5757 \\ \scriptsize(0.5141 : 0.6582)} & \makecell{0.5955 \\ \scriptsize(0.5323 : 0.6421)} & \makecell{0.7109 \\ \scriptsize(0.6650 : 0.7568)} \\
\hline
Titanic & \makecell{0.8163 \\ \scriptsize(0.7942 : 0.8674)} & \makecell{0.7923 \\ \scriptsize(0.7709 : 0.8381)} & \makecell{0.7874 \\ \scriptsize(0.7618 : 0.8335)} & \makecell{0.7861 \\ \scriptsize(0.7568 : 0.8159)} & \makecell{0.5340 \\ \scriptsize(0.5022 : 0.6479)} & \makecell{0.7846 \\ \scriptsize(0.7573 : 0.8289)} & \makecell{0.8076 \\ \scriptsize(0.7860 : 0.8667)} & \makecell{0.8736 \\ \scriptsize(0.8280 : 0.9191)} \\
\hline
\end{tabular}
}
\vspace{0.1cm}
\caption{Median values of AUC and their ranges.}
\label{tab:appendix_median_auc}
\end{table}

\begin{table}[!htbp]
\centering
\begin{tabular}{|l|c|c|c|c|c|c|}
\hline
\textbf{Dataset} & \textbf{100\% Synthetic} & \textbf{1:1} & \textbf{0.1:1} & \textbf{0.01:1} & \textbf{0.001:1} & \textbf{100\% Real Data} \\ \hline
$A_1$ & \makecell{0.6665 \\ \scriptsize(0.6565 : 0.6794)} & \makecell{\textcolor{darkgreen}{0.7008} \\ \scriptsize(0.6674 : 0.7201)} & \makecell{0.6824 \\ \scriptsize(0.6478 : 0.7184)} & \makecell{\textcolor{red}{0.6825} \\ \scriptsize(0.6530 : 0.7056)} & \makecell{\textcolor{blue}{0.6842} \\ \scriptsize(0.6493 : 0.7097)} & \makecell{0.6819 \\ \scriptsize(0.6500 : 0.7174)} \\ \hline
$A_2$ & \makecell{0.7913 \\ \scriptsize(0.7835 : 0.7981)} & \makecell{0.8202 \\ \scriptsize(0.8140 : 0.8276)} & \makecell{0.8243 \\ \scriptsize(0.8164 : 0.8298)} & \makecell{\textcolor{red}{0.8256} \\ \scriptsize(0.8201 : 0.8315)} & \makecell{\textcolor{darkgreen}{0.8266} \\ \scriptsize(0.8204 : 0.8333)} & \makecell{\textcolor{blue}{0.8264} \\ \scriptsize(0.8212 : 0.8323)} \\ \hline
$A_3$ & \makecell{0.6698 \\ \scriptsize(0.6608 : 0.6766)} & \makecell{0.7109 \\ \scriptsize(0.7053 : 0.7175)} & \makecell{\textcolor{red}{0.7180} \\ \scriptsize(0.7124 : 0.7237)} & \makecell{\textcolor{blue}{0.7193} \\ \scriptsize(0.7134 : 0.7250)} & \makecell{0.7179 \\ \scriptsize(0.7115 : 0.7234)} & \makecell{\textcolor{darkgreen}{0.7195} \\ \scriptsize(0.7147 : 0.7246)} \\ \hline
$A_7$ & \makecell{0.8141 \\ \scriptsize(0.7721 : 0.8597)} & \makecell{0.8133 \\ \scriptsize(0.7474 : 0.8607)} & \makecell{0.8232 \\ \scriptsize(0.7775 : 0.8723)} & \makecell{\textcolor{darkgreen}{0.8354} \\ \scriptsize(0.7926 : 0.8904)} & \makecell{\textcolor{red}{0.8327} \\ \scriptsize(0.7864 : 0.8663)} & \makecell{\textcolor{blue}{0.8334} \\ \scriptsize(0.7933 : 0.8821)} \\ \hline
$A_8$ & \makecell{0.7552 \\ \scriptsize(0.7475 : 0.7602)} & \makecell{0.7802 \\ \scriptsize(0.7740 : 0.7849)} & \makecell{\textcolor{darkgreen}{0.7890} \\ \scriptsize(0.7833 : 0.7932)} & \makecell{0.7877 \\ \scriptsize(0.7829 : 0.7949)} & \makecell{\textcolor{red}{0.7879} \\ \scriptsize(0.7829 : 0.7927)} & \makecell{\textcolor{blue}{0.7880} \\ \scriptsize(0.7810 : 0.7941)} \\ \hline
$A_9$ & \makecell{0.6901 \\ \scriptsize(0.6353 : 0.7762)} & \makecell{0.8241 \\ \scriptsize(0.7397 : 0.8807)} & \makecell{\textcolor{blue}{0.8304} \\ \scriptsize(0.7767 : 0.8746)} & \makecell{0.8238 \\ \scriptsize(0.6990 : 0.8802)} & \makecell{\textcolor{red}{0.8239} \\ \scriptsize(0.7391 : 0.8831)} & \makecell{\textcolor{darkgreen}{0.8555} \\ \scriptsize(0.8015 : 0.9038)} \\ \hline
\end{tabular}
\vspace{0.1cm}
\caption{AUC out-of-time validation on 80\% dataset – \textcolor{darkgreen}{Green} indicates the highest value, \textcolor{blue}{Blue} the second highest, and \textcolor{red}{Red} the third highest in each row.}
\label{tab:auc_scores80}
\end{table}

\begin{table}[!htbp]
\centering
\begin{tabular}{|c|c|c|c|c|}
\hline
\textbf{Percent of outliers} & \textbf{Change of AUC} & \textbf{statistic} & \textbf{p-value} & \textbf{Significance, 0.95} \\ \hline
100\% & -0.0158 & 900  & 0.0003  & $\checkmark$ \\ \hline
50\% & -0.0154 & 1103 & 0.0087  & $\checkmark$ \\ \hline
10\% & -0.0059 & 1342 & 0.1337  & $\times$ \\ \hline
7.7\% & 0.0041 & 1471 & 0.3723  & $\times$ \\ \hline
\textcolor{blue}{7.4\%} & \textcolor{blue}{0.0102} & 1408 & 0.2345  & $\times$ \\ \hline
7.1\% & 0.0056 & 1463 & 0.3524  & $\times$ \\ \hline
7\% & 0.0074 & 1342 & 0.1337  & $\times$ \\ \hline
6.9\% & 0.0053 & 1494 & 0.4331  & $\times$ \\ \hline
6.6\% & 0.0058 & 1599 & 0.7722  & $\times$ \\ \hline
6.3\% & -0.0055 & 1545 & 0.5866  & $\times$ \\ \hline
6\% & 0.0076  & 1279 & 0.0725  & $\times$ \\ \hline
\textcolor{darkgreen}{5\%} & \textcolor{darkgreen}{0.0127} & 1221 & 0.0385  & $\checkmark$ \\ \hline
\textcolor{red}{3\%} & \textcolor{red}{0.0096} & 1314 & 0.1028  & $\times$ \\ \hline
1\% & 0.0011 & 1649 & 0.9568  & $\times$ \\ \hline
\end{tabular}
\vspace{0.1cm}
\caption{Change in AUC and Wilcoxon test results of AUC changes for different percentages of outliers for $A_9$ – \textcolor{darkgreen}{Green} indicates the highest value, \textcolor{blue}{Blue} the second highest, and \textcolor{red}{Red} the third highest.}
\label{tab:auc_significant_test}
\end{table}

\begin{figure}[!htbp]
    \centering
    \resizebox{0.8\linewidth}{!}{\includegraphics{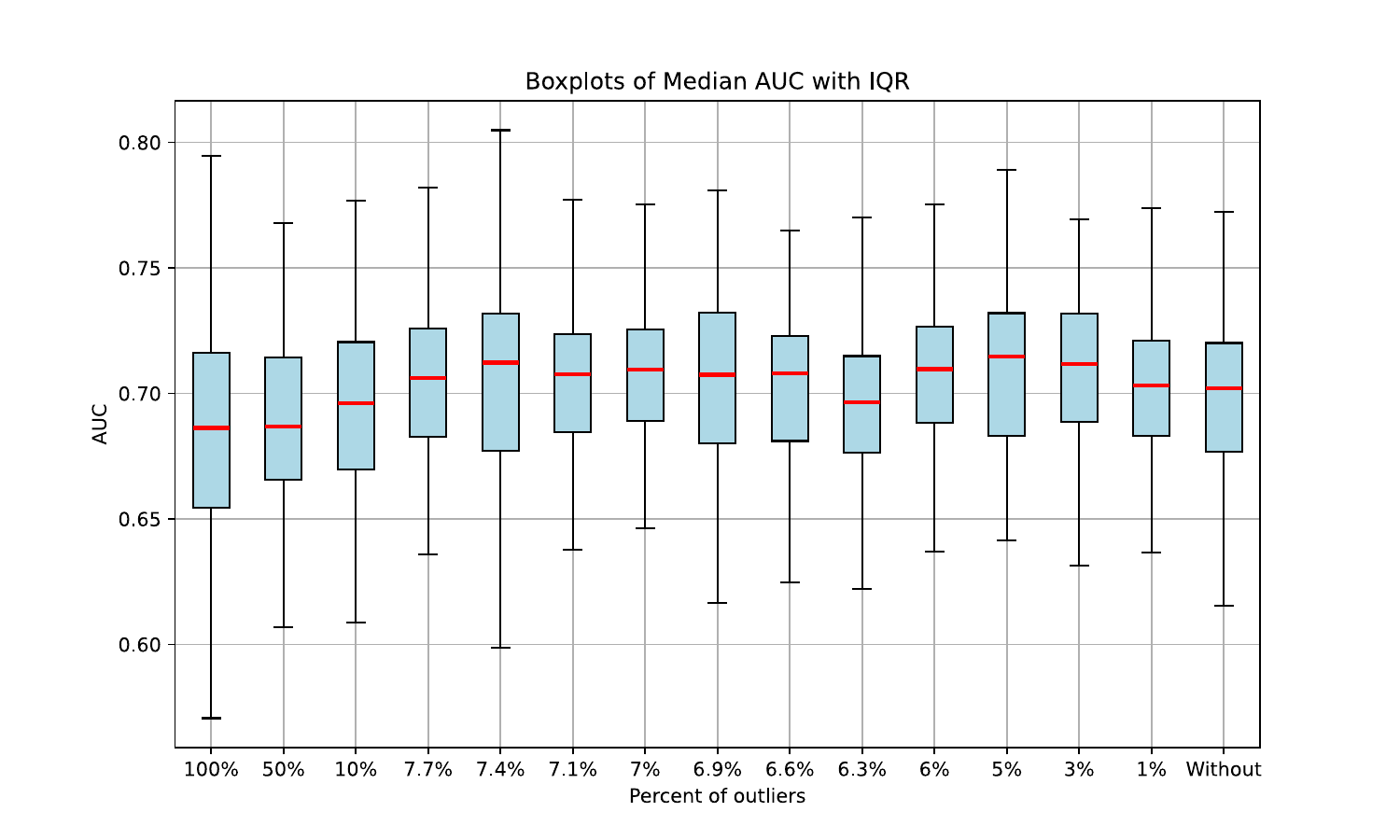}}
    \caption{Median value of AUC and IQR interval for dataset $A_9$.}
    \label{fig:median_auc_outliers}
\end{figure}

\newpage

\section{Appendix: Correlation Analysis}
\label{sec:appendix_correlation_analysis}

\setcounter{figure}{0}
\setcounter{table}{0}

\begin{figure}[!htbp]
    \centering
    \includegraphics[width=\linewidth]{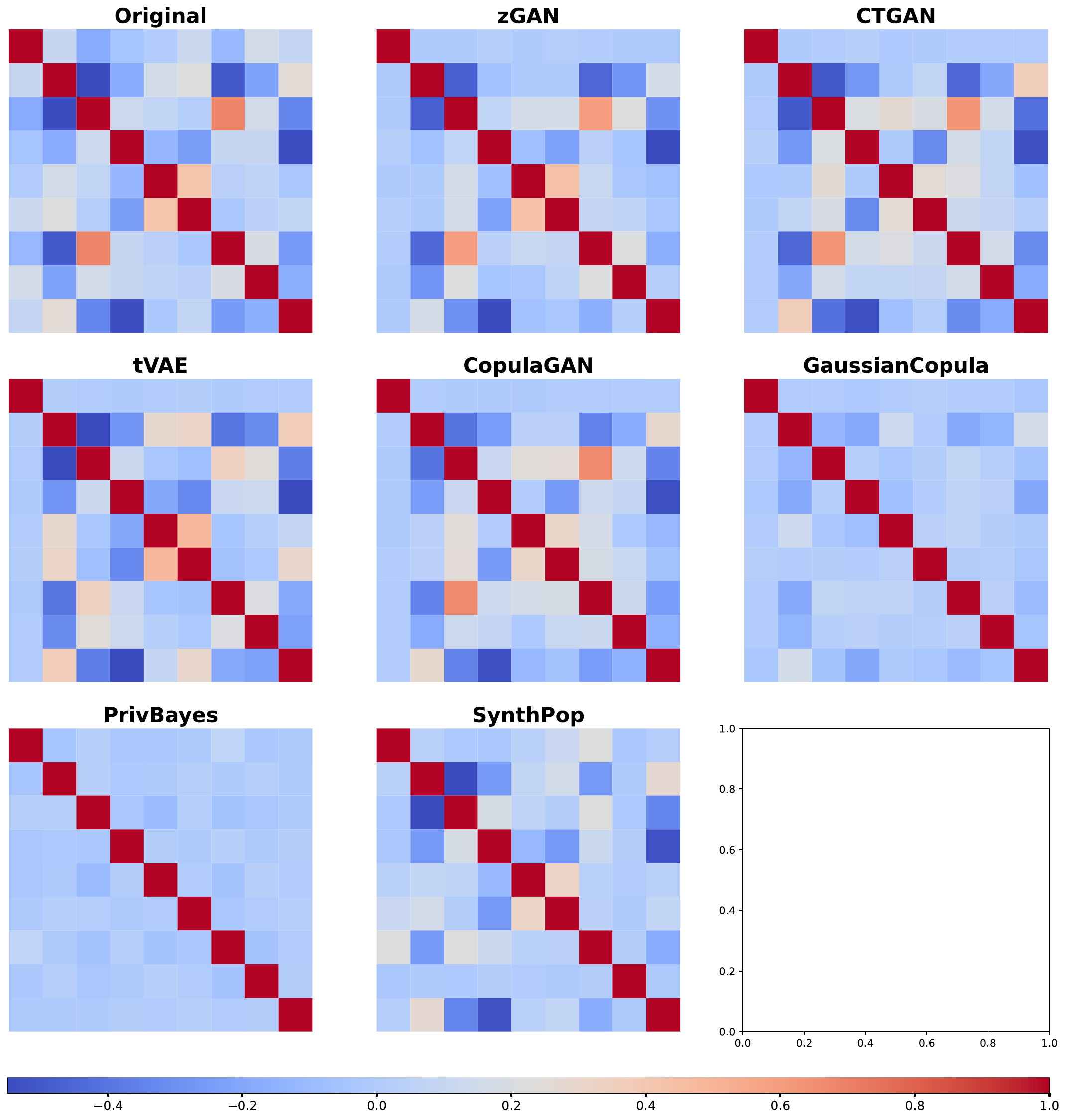}
    \caption{Correlation heatmaps for synthetic and original Titanic datasets.}
    \label{fig:heatmaps}
\end{figure}

\begin{figure}[!htbp]
    \centering
    \includegraphics[width=\linewidth]{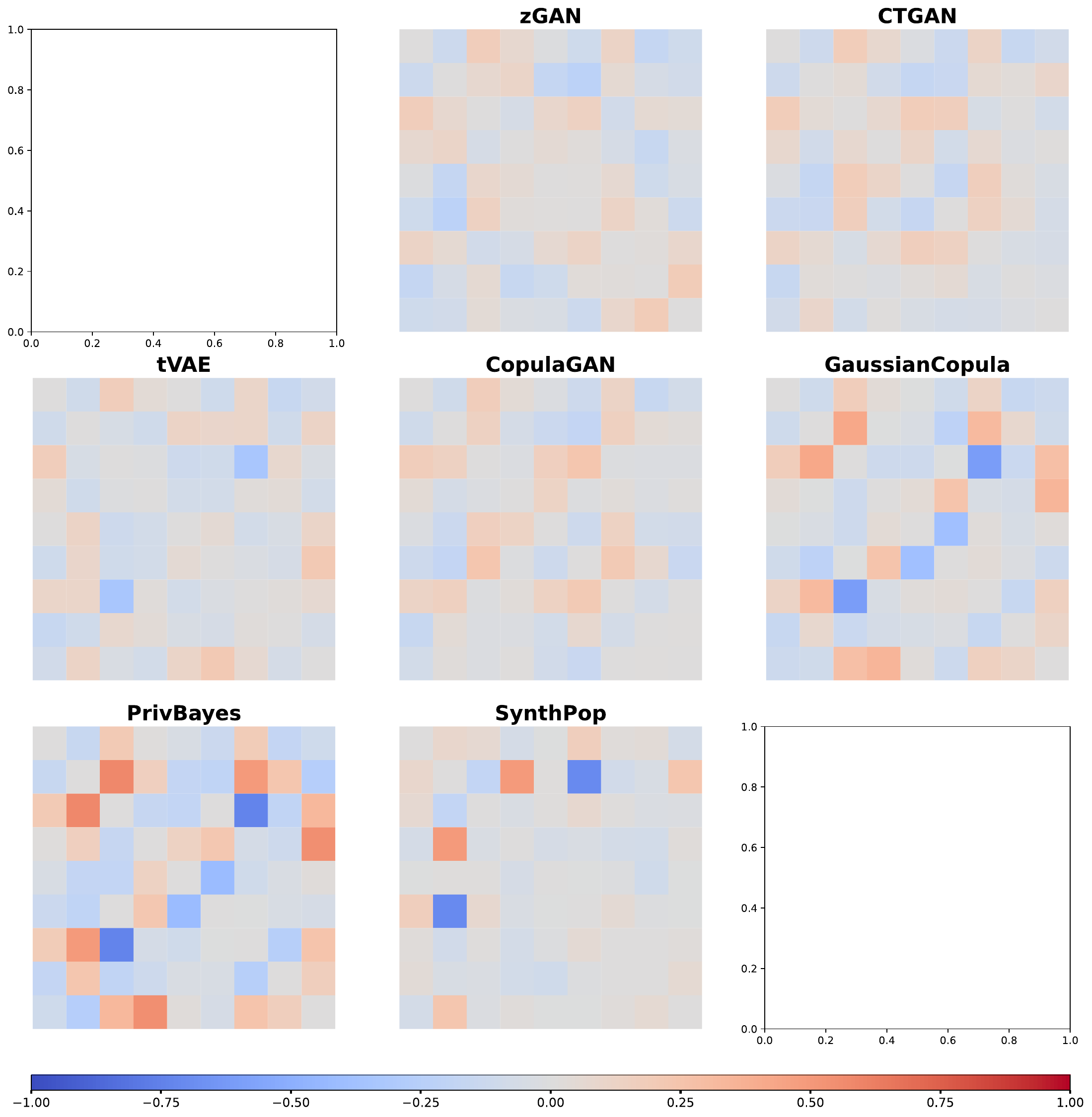}
    \caption{Correlation difference heatmaps for synthetic and original Titanic datasets.}
    \label{fig:difference_heatmaps}
\end{figure}

\end{document}